\pdfoutput=1
\documentclass[10pt,twocolumn,letterpaper]{article}

\usepackage[pagenumbers]{cvpr} %

\usepackage{lipsum}
\usepackage{array}
\usepackage{times}
\usepackage{epsfig}
\usepackage{graphicx}
\usepackage{float}
\usepackage{wrapfig}
\usepackage{amsmath,amssymb,amsthm}
\usepackage{algorithm,algorithmicx,algpseudocode}
\usepackage{bm,xspace}
\usepackage{comment}
\usepackage{multirow}
\usepackage{balance}
\usepackage{url}
\usepackage{booktabs}
\usepackage{etoolbox,siunitx}
\usepackage{calc}
\usepackage{pifont,hologo}
\usepackage{color}
\usepackage{adjustbox}
\usepackage{amsmath}
\usepackage{enumitem}
\usepackage{bbding}  %
\usepackage[normalem]{ulem}  %
\usepackage{contour}
\usepackage[table]{xcolor}
\usepackage{soul}%
\usepackage{tocloft} %
\usepackage{etoc} %
\PassOptionsToPackage{dvipsnames}{xcolor}
\usepackage[pagebackref,breaklinks,colorlinks,linkcolor=link,citecolor=cite]{hyperref}
\usepackage{natbib}
\PassOptionsToPackage{square,numbers,comma,sort}{natbib}

\newcommand{\authorskip}{\hspace{2.5mm}}
\newcommand{\institutionskip}{\hspace{5.0mm}}

\definecolor{blue}{HTML}{004bb3}
\definecolor{red}{HTML}{cc1100}
\definecolor{orange}{HTML}{cc7700}
\definecolor{gray}{HTML}{efefef}
\definecolor{darkgreen}{HTML}{228B22}
\definecolor{darkgray}{HTML}{757575}

\definecolor{cite}{HTML}{3270b5}
\definecolor{link}{HTML}{b53532}
\definecolor{link}{HTML}{cc1100}
\definecolor{scratch}{HTML}{001219}
\definecolor{pretrain}{HTML}{0A9396}

\contourlength{0.8pt}

\newcommand{\figref}[1]{Fig.~\ref{#1}}
\newcommand{\tabref}[1]{Tab.~\ref{#1}}
\newcommand{\secref}[1]{Sec.~\ref{#1}}
\renewcommand{\eqref}[1]{Eq.~\ref{#1}}

\newcolumntype{x}[1]{>{\centering\arraybackslash}p{#1}}
\newcolumntype{y}[1]{>{\raggedright\arraybackslash}p{#1}}
\newcolumntype{z}[1]{>{\raggedleft\arraybackslash}p{#1}}
\newcommand{\tablestyle}[2]{\setlength{\tabcolsep}{#1}\renewcommand{\arraystretch}{#2}\centering\footnotesize}

\DeclareMathSymbol{@}{\mathord}{letters}{"3B}

\newcommand\mypara[1]{\vspace{0mm}\noindent\textbf{#1}}

\makeatletter
\DeclareRobustCommand\onedot{\futurelet\@let@token\@onedot}
\def\@onedot{\ifx\@let@token.\else.\null\fi\xspace}

\newcommand*{\Rom}[1]{\expandafter\@slowromancap\romannumeral #1@}
\newcommand*{\rom}[1]{\expandafter\romannumeral #1}

\def\1{\bm{1}}

\DeclareMathAlphabet{\mathsfit}{\encodingdefault}{\sfdefault}{m}{sl}
\SetMathAlphabet{\mathsfit}{bold}{\encodingdefault}{\sfdefault}{bx}{n}

\let\originalleft\left
\let\originalright\right
\renewcommand{\left}{\mathopen{}\mathclose\bgroup\originalleft}
\renewcommand{\right}{\aftergroup\egroup\originalright}

\def\paperID{200}
\def\confName{CVPR}
\def\confYear{2024}

\begin{document}

\title{\textcolor{blue}{P}oint \textcolor{blue}{T}ransformer \textcolor{red}{V3} Extreme: 1st Place Solution for \\ 2024 Waymo Open Dataset Challenge in Semantic Segmentation}

\author{Xiaoyang Wu\textsuperscript{\mdseries1,2} \authorskip 
Xiang Xu\textsuperscript{\mdseries2} \authorskip
Lingdong Kong\textsuperscript{\mdseries3} \\
Liang Pan\textsuperscript{\mdseries4} \authorskip
Ziwei Liu\textsuperscript{\mdseries4} \authorskip 
Tong He\textsuperscript{\mdseries2} \authorskip
Wanli Ouyang\textsuperscript{\mdseries2} \authorskip
Hengshuang Zhao\textsuperscript{\mdseries1}
 \\ \\
\textsuperscript{1}HKU \institutionskip
\textsuperscript{2}SH AI Lab \institutionskip
\textsuperscript{3}NUS \institutionskip
\textsuperscript{4}NTU \\ 
{\tt\small \url{https://github.com/Pointcept/Pointcept}}
}

\maketitle
\begin{abstract}
\vspace{-4.5mm}

In this technical report, we detail our first-place solution for the 2024 Waymo Open Dataset Challenge's semantic segmentation track. We significantly enhanced the performance of Point Transformer V3 on the Waymo benchmark by implementing cutting-edge, plug-and-play training and inference technologies. Notably, our advanced version, Point Transformer V3 Extreme, leverages multi-frame training and a no-clipping-point policy, achieving substantial gains over the original PTv3 performance. Additionally, employing a straightforward model ensemble strategy further boosted our results. This approach secured us the top position on the Waymo Open Dataset semantic segmentation leaderboard, markedly outperforming other entries.

\vspace{-4.5mm}
\end{abstract}

\section{Introduction}
\label{sec:intro}

\begin{table}[!t]
    \begin{minipage}{0.48\textwidth}
    \centering
        \vspace{-2mm}
        \tablestyle{2pt}{1.08}
        \begin{tabular}{lclc}\toprule
\multicolumn{2}{c}{Original} &\multicolumn{2}{c}{\cellcolor[HTML]{efefef}Extreme} \\
\cmidrule(lr){1-2} \cmidrule(lr){3-4}
Config &Value &Config &Value \\\midrule
optimizer &AdamW &optimizer &AdamW \\
scheduler &Cosine &scheduler &Cosine \\
criteria &CrossEntropy~(1) &criteria &CrossEntropy~(1) \\
&Lovasz~\cite{berman2018lovasz}~(1) & &Lovasz~\cite{berman2018lovasz}~(1) \\
learning rate &2e-3 &learning rate &2e-3 \\
block lr scaler &1e-1 &block lr scaler &1e-1 \\
weight decay &5e-3 &weight decay &5e-3 \\
batch size &12 &batch size &12 \\
datasets &Waymo &datasets &Waymo \\
warmup epochs &2 &warmup epochs &2 \\
epochs &50 &epochs &50 \\
\cellcolor[HTML]{efefef}frames & \cellcolor[HTML]{efefef}[0] & \cellcolor[HTML]{efefef}frames & \cellcolor[HTML]{efefef}[0, -1, -2] \\
\cellcolor[HTML]{efefef}model ensemble & \cellcolor[HTML]{efefef}$\times$ & \cellcolor[HTML]{efefef}model ensemble & \cellcolor[HTML]{efefef}\checkmark \\
\bottomrule
\end{tabular}

        \vspace{-3mm}
        \caption{\textbf{Training settings.} }\label{tab:waymo_training_settings}
        \vspace{1mm}
    \end{minipage}
    \begin{minipage}{0.48\textwidth}
    \centering
        \tablestyle{0.7pt}{1.07}
        \begin{tabular}{llcc}\toprule
Augmentations &Parameters &Original &\cellcolor[HTML]{efefef}Extreme \\\midrule
random rotate &axis: z, angle: [-1, 1], p: 0.5 &\checkmark &\checkmark \\
\cellcolor[HTML]{efefef}point clip & \cellcolor[HTML]{efefef}range: [-75.2, -75.2, -4, 75.2, 75.2, 2] &\cellcolor[HTML]{efefef}\checkmark &\cellcolor[HTML]{efefef}$\times$ \\
random scale &scale: [0.9, 1.1] &\checkmark &\checkmark \\
random flip &p: 0.5 &\checkmark &\checkmark \\
random jitter &sigma: 0.005, clip: 0.02 &\checkmark &\checkmark \\
grid sampling &grid size: 0.05 &\checkmark &\checkmark \\
\bottomrule
\end{tabular}
        \vspace{-3mm}
        \caption{\textbf{Data augmentations.} }\label{tab:waymo_data_augmentations}
        \vspace{-9mm}
    \end{minipage}
\end{table}

In recent years, the Waymo Open Dataset Challenge~\cite{sun2020waymo} has emerged as a premier arena for showcasing advancements in autonomous driving technologies. The 2024 iteration of this challenge continued to push the boundaries of what is achievable in 3D perception, leveraging the rich and diverse data provided by Waymo. The Waymo Open Dataset is characterized by its high-resolution LiDAR scans and comprehensive annotations, making it ideal for developing and testing cutting-edge 3D perception algorithms~\cite{spconv2022, pointcept2023, wu2024ppt, wu2023masked, peng2024oacnns}. This technical report presents our winning entry for the 2024 Waymo Open Dataset Challenges semantic segmentation track. Our approach builds upon the foundation of the Point Transformer V3 (PTv3)~\cite{wu2024ptv3}, known for its robustness and efficiency in handling point cloud data. We optimized PTv3 for the specific challenges posed by the Waymo dataset, implementing several plug-and-play training and inference technologies that significantly enhance performance.

Key to our strategy was the implementation of multi-frame training, which incorporates data from two previous frames to enrich the perception of current LiDAR frames. This technique, combined with a no-clipping-point policy that avoids discarding data points outside a specified range, provided a deeper insight into the spatial and temporal aspects of the dataset. This enhanced version, termed Point Transformer V3 Extreme, achieved substantial performance improvements over the original PTv3 metrics reported in earlier works. Furthermore, by incorporating a simple yet effective model ensemble strategy, we were able to achieve unprecedented accuracy, securing the first-place position on the semantic segmentation leaderboard. The detailed parameter settings are presented in \tabref{tab:waymo_training_settings} and \tabref{tab:waymo_data_augmentations}. Our methods outperformed other competitive entries by remarkable margins, demonstrating the potential of advanced transformer architectures in complex, real-world environments like those represented in the Waymo Open Dataset.

\begin{figure*}[t]
    \vspace{-1.5mm}
    \centering
    \includegraphics[width=\linewidth]{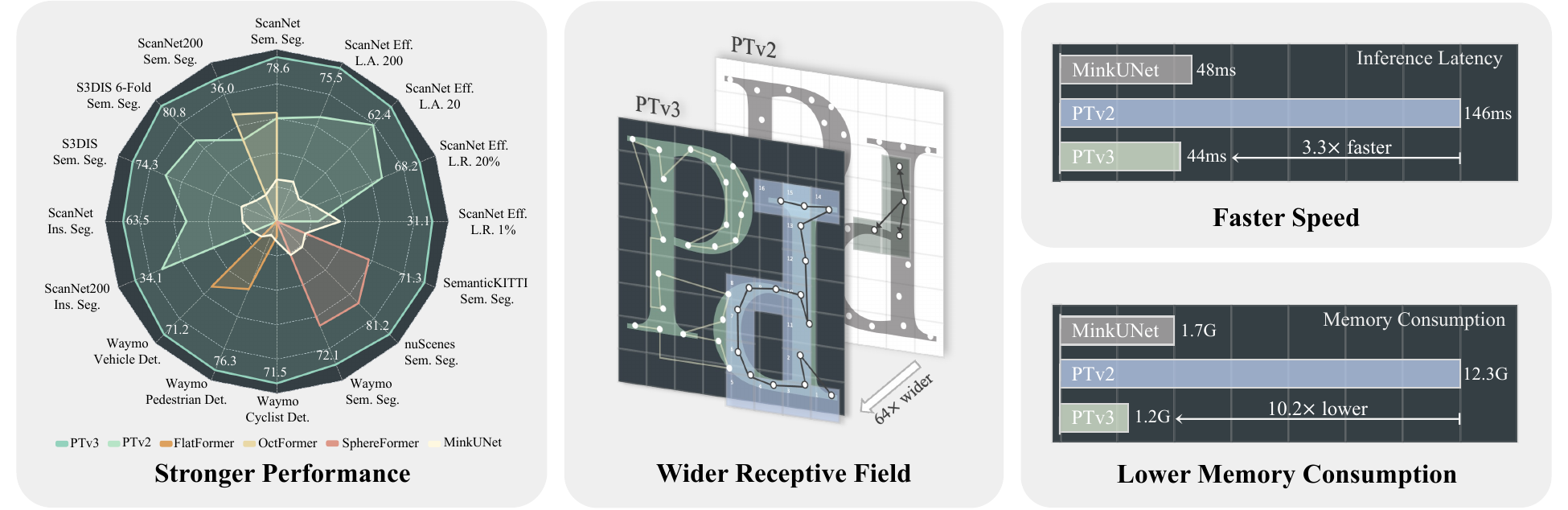}
    \vspace{-6.5mm}
    \caption{\textbf{Overview of Point Transformer V3 (PTv3).} Compared to its predecessor, PTv2~\cite{wu2022point}, our PTv3 shows superiority in the following aspects: 1. \textit{Stronger performance.}  PTv3 achieves state-of-the-art results across a variety of indoor and outdoor 3D perception tasks. 2. \textit{Wider receptive field.} Benefit from the simplicity and efficiency, PTv3 expands the receptive field from 16 to 1024 points. 3. \textit{Faster speed.} PTv3 significantly increases processing speed, making it suitable for latency-sensitive applications. 4. \textit{Lower Memory Consumption.} PTv3 reduces memory usage, enhancing accessibility for broader situations.}
    \label{fig:teaser}
    \vspace{-4mm}
\end{figure*}

\section{Point Transformer V3 Extreme}
\label{sec:method}
This section first has a revisit of the Point Transformer V3 in \secref{sec:revisit}. After that, we go through the details of additional training technologies in \secref{sec:technologies}. The detailed parameter settings are presented in \tabref{tab:waymo_training_settings} and \tabref{tab:waymo_data_augmentations}.

\subsection{Revisit Point Transformer V3}
\label{sec:revisit}

\mypara{Scaling principle.} Enhanced with large-scale pre-training, SparseUNet~\cite{choy20194d} surpasses Point Transformers~\cite{zhao2021point,wu2022point} in accuracy while remaining efficient. Yet, Point Transformers fails to scale up due to limitations in efficiency, which inspired the hypothesis that model performance is more significantly influenced by scale than by complex design details. Backbone design should prioritize simplicity and efficiency over the accuracy of certain mechanisms. Efficiency enables scalability, which further brings a stronger accuracy.

\mypara{Breaking the curse of permutation invariance} Classical point cloud transformers build upon point-based backbones~\cite{qi2017pointnet, qi2017pointnet++}, which treat point clouds as unstructured data and rely on neighboring query algorithms like the k-nearest neighbor (kNN). Yet kNN is extremely inefficient due to the difficulty in parallelization, which further raises the question of whether we really need the accurate neighbours queried by kNN. Considering that attention is adaptive to kernel shape, it is worth trading the accurate spatial proximity for additional scalability. Inspired by OctFormer~\cite{Wang2023OctFormer} and FlatFormer~\cite{liu2023flatformer}, PTv3 abandoned the unstructured nature of the point cloud, exploring a strategy to turn unstructured sparse data into structured 1D data as language tokens while preserving necessary spatial proximity to attention.

\begin{figure}[t]
    \vspace{-1mm}
    \centering
    \includegraphics[width=\linewidth]{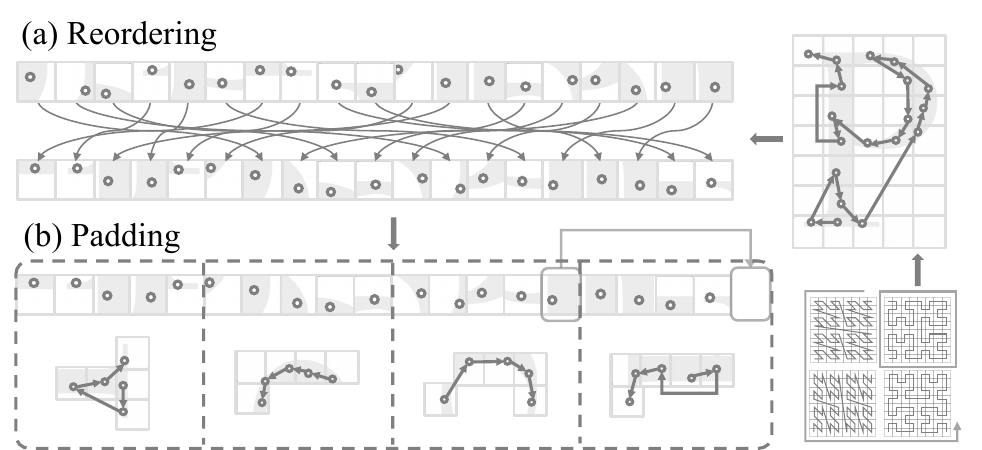}
    \vspace{-7mm}
    \caption{\textbf{Patch grouping.} (a) Reordering point cloud according to order derived from a specific serialization pattern. (b) Padding point cloud sequence by borrowing points from neighboring patches to ensure it is divisible by the designated patch size.}
    \label{fig:patch_grouping}
    \vspace{-1mm}
    \centering
    \includegraphics[width=\linewidth]{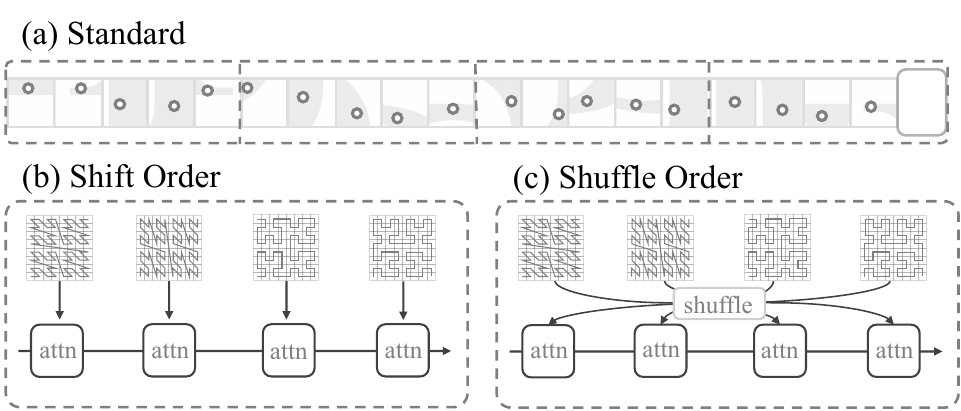}
    \vspace{-7mm}
    \caption{\textbf{Patch interaction.} (a) Standard patch grouping with a regular, non-shifted arrangement; (d) Shift Order where different serialization patterns are cyclically assigned to successive attention layers; (d) Shuffle Order, where the sequence of serialization patterns is randomized before being fed to attention layers.}
    \label{fig:patch_interaction}
    \vspace{-7mm}
\end{figure}

\begin{figure}[t]
    \vspace{-1.1mm}
    \centering
    \includegraphics[width=\linewidth]{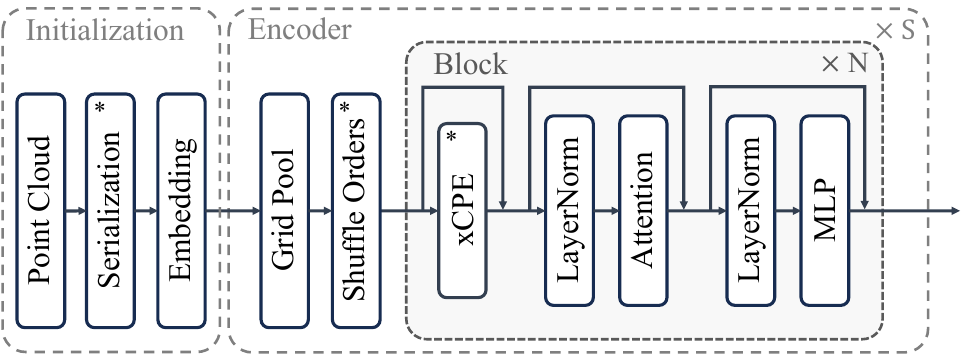}
    \vspace{-7mm}
    \caption{\textbf{Overall architecture.}}
    \label{fig:architecture}
    \vspace{-7mm}
\end{figure}

\mypara{Serialization \& attention.} Space-filling curves are paths that traverse every point within a high-dimensional discrete space, preserving spatial proximity to a certain extent. The serialization of point clouds involves sorting points according to the traversal order defined by a specific space-filling curve. This ordering effectively rearranges the points in a way that respects the spatial organization dictated by the curve, ensuring that neighboring points in the data structure are also spatially close. By reordering point clouds through serialization and incorporating necessary padding operations, the unordered point cloud is transformed into a structured format (see \figref{fig:patch_grouping}). Consequently, attention mechanisms optimized for structured data can be effectively applied to these serialized point clouds. To optimize performance across various benchmarks, PTv3 employs both local attention~\cite{liu2021Swin} and flash attention~\cite{dao2022flashattention, dao2023flashattention2}. For local attention, PTv3 facilitates patch interaction by utilizing various serialization patterns across different attention layers (see \figref{fig:patch_interaction}). Additionally, PTv3 adopts a sparse convolution layer, prepended with a skip connection, as conditional positional encoding~\cite{Wang2023OctFormer, chu2021cpe}, named xCPE. The overall model architecture is visualized in \figref{fig:architecture}.

\subsection{Training Technologies}
\label{sec:technologies}

\begin{figure}[t]
    \centering
    \includegraphics[width=\linewidth]{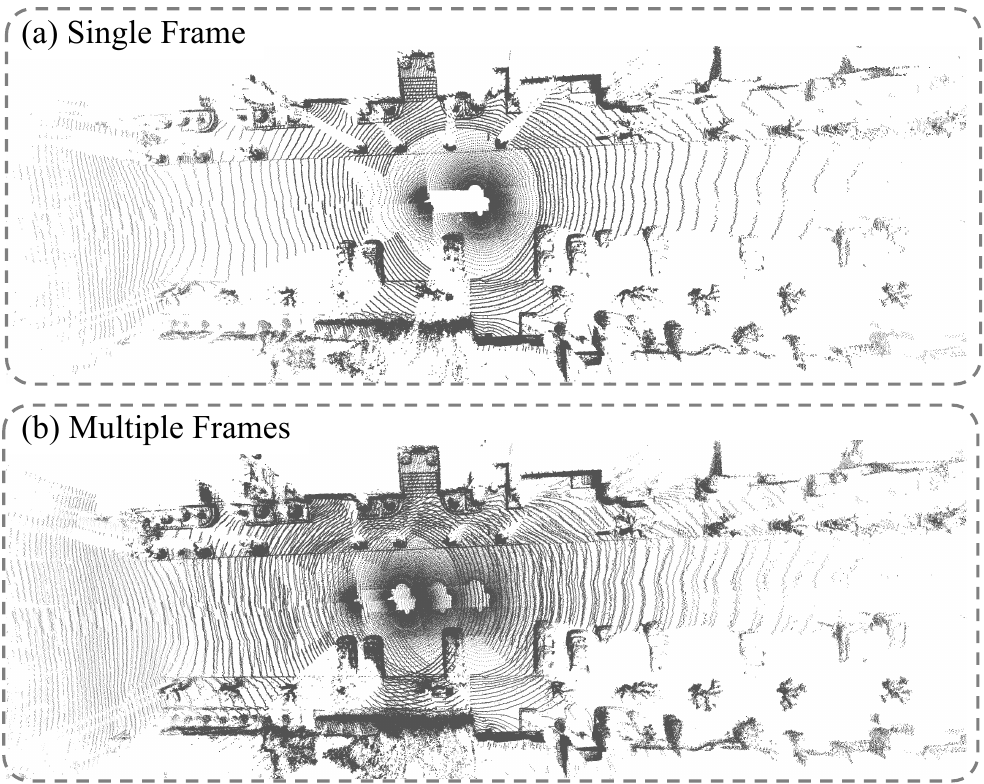}
    \vspace{-7mm}
    \caption{\textbf{Visualization of Multi-frames Concatenation.}}
    \label{fig:frames}
    \vspace{-4mm}
\end{figure}

\mypara{Multi-frame training.} Perceiving distant ranges in a LiDAR point cloud, far from the center, is challenging due to insufficient sampling. An intuitive solution is to concatenate past LiDAR frames with the current frame after a coordinate alignment to supplement the less-sampled areas (see \figref{fig:frames}). Specifically, we incorporate two past labelled frames as additional references during both our training and inference processes, utilizing all of them for supervision during training for convenience.

\mypara{Non-clipping proxy.} However, merely enabling multi-frame training does not automatically result in significant enhancements for perception tasks. We have found that the full potential of multi-frame training is unlocked only when it is combined with a non-clipping strategy. Traditionally, clipping points to a specific range, such as [-75.2, -75.2, -4, 75.2, 75.2, 2] for the Waymo Open Dataset, was a necessary preprocessing step for perception tasks in outdoor scenarios. This was largely because the perception systems~\cite{shi2022pillarnet,fan2022embracing,yin2021center} for autonomous driving, which often rely on submanifold sparse convolution, struggle to effectively incorporate isolated points that frequently occur at distant ranges in open-space LiDAR point clouds. Unlike these systems, PTv3, which organizes point clouds into a structured 1D array, does not suffer from this disadvantage. Without the limitations imposed by a clipping proxy, PTv3 effectively leverages additional information from past frames, which significantly enhances the semantic segmentation mIoU on the Waymo Open Dataset validation split from 72.1\% to 74.8\%.

\begin{table}[!t]
    \begin{minipage}{0.48\textwidth}
    \centering
        \tablestyle{7pt}{1.08}
        \begin{tabular}{lcccc}\toprule
Sem. Seg.&\multicolumn{2}{c}{PTv3~\cite{wu2024ptv3}} &\multicolumn{2}{c}{PTv3-EX} \\\cmidrule(lr){2-3}\cmidrule(lr){4-5}
&val &test &val &test \\
Model Ensemble &- &\checkmark &- &\checkmark \\
Params. & 46.2M & 46.2M$\times$3 & 46.2M & 46.2M$\times$3\\
Training Latency & 245ms & 245ms$\times$3 & 482ms & 482ms$\times$3\\
Inference Latency & 132ms & 132ms$\times$3 & 253ms & 253ms$\times$3 \\
\midrule
Car &0.9447 &0.9571 &0.9463 &0.9662 \\
Truck &0.6207 &0.6793 &0.6283 &0.7397 \\
Bus &0.8665 &0.7482 &0.8920 &0.7792 \\
Other Vehicle &0.3582 &0.3654 &0.4857 &0.3681 \\
Motorcyclist &0.1630 &0.0000 &0.3946 &0.1514 \\
Bicyclist &0.7878 &0.9010 &0.8030 &0.9203 \\
Pedestrian &0.9120 &0.9264 &0.9162 &0.9372 \\
Sign &0.7235 &0.7404 &0.7664 &0.7502 \\
Traffic Light &0.3607 &0.3373 &0.4276 &0.3465 \\
Pole &0.7778 &0.8157 &0.8036 &0.8254 \\
Construction Cone &0.7562 &0.6690 &0.7405 &0.6693 \\
Bicycle &0.7821 &0.6851 &0.7772 &0.7226 \\
Motorcycle &0.9034 &0.8070 &0.9154 &0.8263 \\
Building &0.9606 &0.9736 &0.9636 &0.9751 \\
Vegetation &0.9189 &0.8812 &0.9242 &0.8901 \\
Tree Trunk &0.6860 &0.7500 &0.7069 &0.7575 \\
Curb &0.7152 &0.7520 &0.7226 &0.7648 \\
Road &0.9348 &0.9306 &0.9368 &0.9330 \\
Lane Marker &0.5712 &0.4967 &0.5726 &0.5111 \\
Other Ground &0.5206 &0.5255 &0.5248 &0.5414 \\
Walkable &0.8167 &0.7357 &0.8196 &0.7538 \\
Sidewalk &0.7872 &0.8733 &0.7891 &0.8788 \\  \midrule
 \cellcolor[HTML]{efefef}mIoU & \cellcolor[HTML]{efefef}0.7213 & \cellcolor[HTML]{efefef}0.7068 & \cellcolor[HTML]{efefef}\textbf{0.7480} & \cellcolor[HTML]{efefef}\textbf{0.7276} \\
\bottomrule
\end{tabular}
        \vspace{-3mm}
        \caption{\textbf{Results on Waymo Open Dataset.} Latency and memory usage were assessed on a single RTX 4090 GPU, with the batch size fixed at 1 and models are trained with 4 NVIDIA a100 GPUs.}\label{tab:waymo_sem_seg}
        \vspace{-7mm}
    \end{minipage}
\end{table}

\mypara{Model ensemble.} One technique that consistently boosts model performance is model ensembling. In our approach, we independently train three PTv3 models and combine their predicted logits to form our final submission. It's important to note that we discourage using this technique for performance comparisons, especially on the validation split, as it can lead to unfair comparisons. We have limited the use of this technology to the Waymo Challenge test split. We also advise future researchers to refrain from using this technique for validation comparisons.

\section{Conclusion}
\label{sec:conclusion}

Enhanced with multi-frame training, a non-clipping strategy, and model ensembling, we have significantly extended the capabilities of Point Transformer V3. Specifically, on the Waymo Open Dataset, the validation mIoU increased from 72.1\% to 74.8\%, and the test mIoU rose from 70.7\% to 72.8\% (details provided in \tabref{tab:waymo_sem_seg}). We hope these technologies and results will inspire future research.

{
\small
\bibliographystyle{ieeenat_fullname}
\bibliography{main}
}

\end{document}